\def\year{2020}\relax
\documentclass[letterpaper]{article} 
\usepackage{aaai20}  
\usepackage{times}
\usepackage{helvet}
\usepackage{courier}
\usepackage{soul}
\usepackage{url}
\usepackage[utf8]{inputenc}
\usepackage{graphicx}
\usepackage{amssymb,amsmath}
\usepackage{booktabs}
\usepackage{algorithm}
\usepackage{algorithmic}
\usepackage{courier}
\usepackage{comment}
\usepackage{amssymb,amsmath}
\usepackage{multirow}
\usepackage{subcaption}
\usepackage{enumerate}
\title{Multi-View Multiple Clusterings using Deep Matrix Factorization}
\author{
Shaowei Wei,\textsuperscript{\rm 1}
Jun Wang,\textsuperscript{\rm 1,}\thanks{Corresponding author, kingjun@swu.edu.cn (Jun Wang).}
Guoxian Yu,\textsuperscript{\rm 1,2}
Carlotta Domeniconi,\textsuperscript{\rm 3}
Xiangliang Zhang\textsuperscript{\rm 2}\\
\textsuperscript{\rm 1}College of Computer and Information Sciences, Southwest University, Chongqing, China\\
\textsuperscript{\rm 2}CEMSE, King Abdullah University of Science and Technology, Thuwal, SA\\
\textsuperscript{\rm 3}Department of Computer Science, George Mason University, VA, USA\\
\{swwei2019, kingjun, gxyu\}@swu.edu.cn,
carlotta@cs.gmu.edu,
xiangliang.zhang@kaust.edu.sa
}

\begin{document}
\maketitle
\begin{abstract}
Multi-view clustering aims at integrating complementary information from multiple heterogeneous views to improve clustering results. Existing multi-view clustering solutions can \emph{only} output a single clustering of the data. Due to their multiplicity, multi-view data, can have different \emph{groupings} that are reasonable and interesting from different perspectives. However, how to find multiple, meaningful, and diverse clustering results from multi-view data is still a rarely studied and challenging topic in multi-view clustering and multiple clusterings. In this paper, we introduce a deep matrix factorization based solution (DMClusts) to discover multiple clusterings.  DMClusts gradually factorizes  multi-view data matrices into representational subspaces layer-by-layer and generates one clustering in each layer. To enforce the diversity between generated clusterings, it minimizes a new redundancy quantification term derived from the proximity between samples in these subspaces. We further introduce an iterative optimization procedure to simultaneously seek multiple clusterings with quality and diversity. Experimental results on benchmark datasets confirm that DMClusts outperforms state-of-the-art multiple clustering solutions.
\end{abstract}


\section{Introduction}
Many real-world data include diverse types of feature views. For example, web images have both visual and textual features; a protein  has structure and interactome features. The various feature views embody consistent and complementary information of the same objects, and have produced intensive research in multi-view learning \cite{bickel2004multi,zhao2017mvoverview}. The fusion of feature views enables not only the achievement of a comprehensive composite view of the objects, but also facilitates the associated learning task  \cite{nie2017multi,tan2018incomplete}.

Various efforts have been focused on the development of effective multi-view clustering (MVC) algorithms. Some methods achieve  clustering by co-regularization \cite{kumar2011co,cheng2013flexible}, correlation analysis \cite{chaudhuri2009multi}, or multiple kernel learning \cite{gonen2011multiple,liu2019multiple};  other approaches learn the shared subspace to extract complementary and shared information of multi-view data, and perform clustering therein \cite{li2014partial,gao2015multi,zhao2017dmf,zong2017mvnmf,kang2019aligned}.

\begin{figure}[t]
  \centering
  \includegraphics[width=.8\columnwidth]{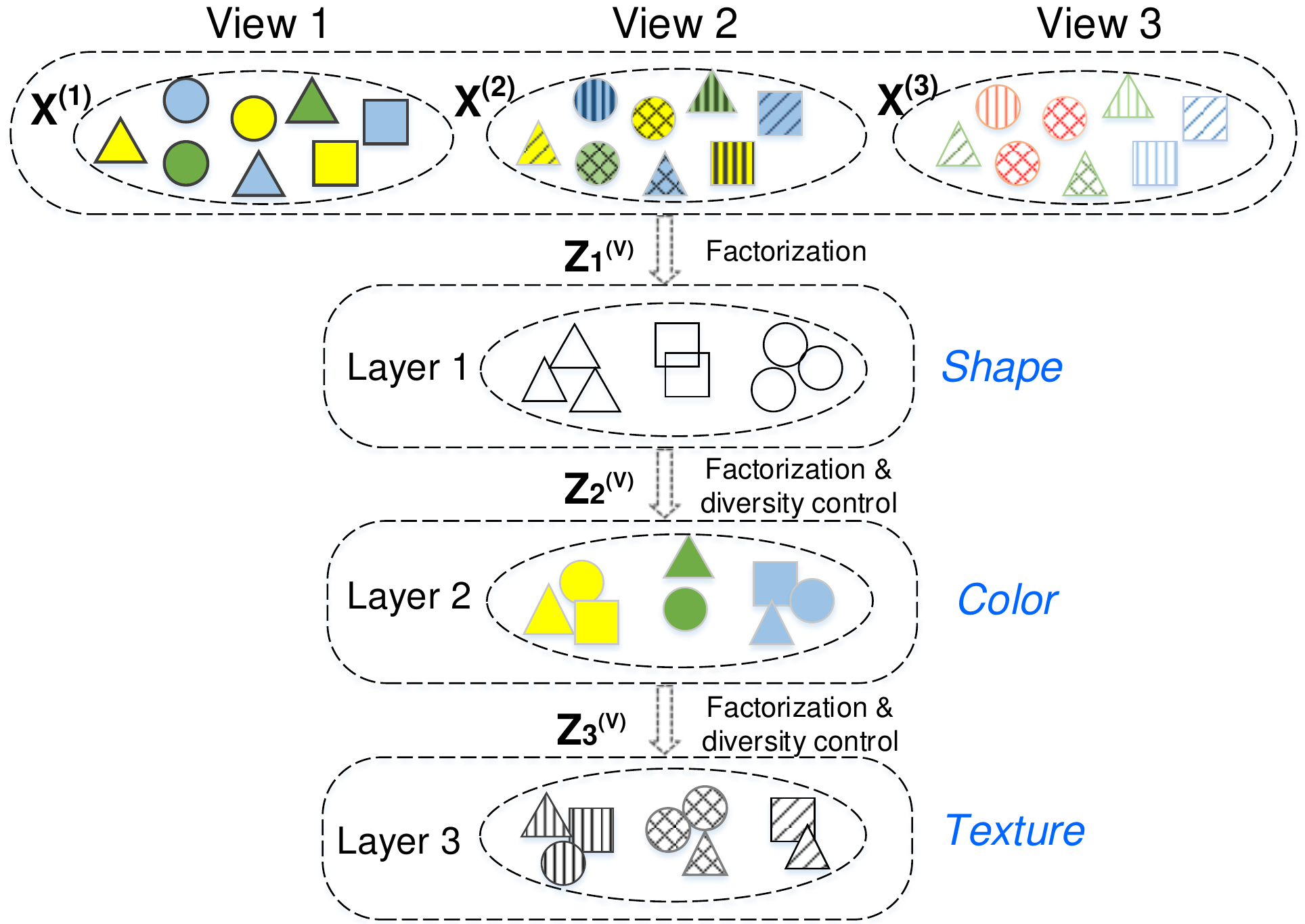}
  \caption{An example of grouping the same objects with three views via deep matrix factorization and diversity control layer-by-layer. The {\it Shape} clustering is generated from all three views, while the {\it Color} and {\it Texture} clusterings are generated from the first two views and the last two views, respectively.}
  \label{fig:case}
\end{figure}

Existing MVC solutions  focus on generating a single clustering; they fail to present  different but meaningful clusterings of the same multi-view data \cite{fanaee2018multi}. For example, the three-view objects in Figure \ref{fig:case} have different shapes, colors, and textures. The aforementioned MVC solutions group these objects mainly by shape. But they can also be clustered according to the  shared color and texture. These groupings are meaningful but different. In other words, multiple clustering is concerned with both the \emph{quality} and \emph{diversity} of alternative clusterings. Although multiple clusterings can present alternative and overlooked meaningful clusterings of the same objects, it is a known dilemma to balance  diversity and quality \cite{bailey2013alternative}. Given this challenge, a number of solutions have been introduced to generate alternative clusterings in different subspaces \cite{cui2007OSC,mautz2018NrKMeans,wang2019MISC}, by meta clustering of base clusterings \cite{caruana2006meta}, by referring to already explored clusterings \cite{bae2006coala,yang2017MNMF}, or by simultaneously reducing the redundancy between clusterings \cite{wang2018mcc,yao2019mccss}. However, they still focus on \emph{single-view} data. One naive extension is to concatenate diverse feature vectors of the same objects into a longer one, and then directly apply  off-the-shelf multiple clustering solutions on concatenated vectors. However, this concatenation overrides the intrinsic nature of multi-view data, and thus reduces the quality and increases the redundancy of explored clusterings, as our experiments will show.

To find multiple clusterings on multi-view data, \cite{yao2019MVMC} recently proposed a solution called multi-view multiple clustering (MVMC). MVMC extracts the individual and shared similarity matrices of multi-view data based on the adapted self-representation learning \cite{luo2018consistent}, and then applies semi-nonnegative matrix factorization \cite{ding2010convex} on each combination of the individual and common similarity data matrices to generate alternative clusterings, where the quality is pursued by the commonality matrix and the diversity is obtained by the individuality matrix. However, MVMC:  (a) does not differentiate the relevance of different views and suffers from low-quality (irrelevant) data views; (b) does not maintain well the quality and diversity of multiple clusterings; (c) cannot be applied for datasets with a large number of samples, since it has to factorize the combined similarity matrix with size equal to the number of samples.

In this paper, we introduce a deep matrix factorization based solution (DMClusts, as illustrated in Figure \ref{fig:case}) to generate multiple diverse clusterings of good quality in a layer-wise fashion. DMClusts collaboratively factorizes the multi-view data matrices into multiple representational subspaces layer-by-layer, and seeks an alternative clustering of quality per layer. To achieve  diversity among the clusterings, it reduces their redundancy by means of a new balanced redundancy quantification term, which jointly considers the case when two objects are often grouped together and the case when they are in different clusters of the subspaces. We further introduce an iterative optimization procedure to simultaneously seek multiple clusterings in a layer-wise fashion. The main contributions of our work are:
\begin{enumerate}[(i)]

\item We introduce a deep matrix factorization based solution (DMClusts) to seek multiple clusterings by fusing the consensus and complementary information of multi-view data, and by enforcing the diversity between the clusterings layer-by-layer. DMClusts  can credit different degrees of relevance to different views; as such, it's less sensitive to noisy (or low-quality) ones.

\item DMClusts introduces a balanced redundancy quantification term, which jointly considers the case that two samples are often nearby in the representational subspace per layer, and the reverse case that they are often faraway per layer, to comprehensively quantify the redundancy of multiple clusterings, whilst existing similar quantification overlooks the latter case. Extensive experiments on benchmark datasets show that DMClusts  significantly outperforms other related competitive multiple clusterings solutions \cite{yao2019MVMC,wang2019MISC,yang2017MNMF,ye2016generalized,jain2008deckmeans,cui2007OSC} and the deep matrix factorization \cite{trigeorgis2017deep} in finding multiple clusterings with quality and diversity.
\end{enumerate}


\section{Our Method}
\subsection{Overview of deep matrix factorization}
Matrix factorization techniques have been extensively adopted for data analysis and representation learning in various domains \cite{tang2017trimf,fu2018mflda,li2019deepmf}. For example, NMF (nonnegative matrix factorization) \cite{lee2001NMF} can decompose a nonnegative data matrix $\mathbf{X}$ into two factor matrices $\mathbf{X}\approx \mathbf{Z} \mathbf{H}$, the nonnegative constraints imposed on factors allow for better interpretability and lead to significantly growing application of  NMF  and its variants \cite{ding2010convex,cai2011graph,vzitnik2014dfmf}. By taking $\mathbf{Z} \in \mathbb{R}^{d\times K}$ as $K$ cluster centroids in the $d$-dimensional feature space, and $\mathbf{H} \in \mathbf{R}^{K \times n}$ as the soft membership indicators of $n$ samples to these centroids, semi-NMF \cite{ding2010convex} is equivalent to a soft version of $k$-mean clustering.  To absorb mix-sign $\mathbf{X}$, semi-NMF only imposes the nonnegative constraints on $\mathbf{H}$.

To explore the complex hierarchical structure and to eliminate noise in the  data matrix $\mathbf{X}$ with different modalities, and motivated by the idea and robustness of deep representation learning \cite{hinton2006nndr,bengio2009deeplearning}, \cite{trigeorgis2017deep} extends semi-NMF to deep semi-NMF (DMF) as follows:
\begin{equation}
\small
\centering
\begin{split}
\mathbf{X}&\approx \mathbf{Z}_1 \mathbf{H}_1\\
\mathbf{X}&\approx \mathbf{Z}_1 \mathbf{Z}_2 \mathbf{H}_2 \\
&\cdots \\
\mathbf{X}&\approx \mathbf{Z}_1 \mathbf{Z}_2 ... \mathbf{Z}_m \mathbf{H}_m
\end{split}
\label{Eq1}
\end{equation}
where $\mathbf{Z}_l \in \mathbf{R}^{K_{l-1} \times K_l}$ is the $l$-th ($l \leq m$) layer basis matrix, and $\mathbf{H}_l \in \mathbb{R}^{K_l \times n} (\geq 0)$ is the $l$-th layer representation matrix. By taking $\mathbf{Z}_1, \cdots, (\mathbf{Z}_1\mathbf{Z}_2 \cdot \mathbf{Z}_l) \in \mathbf{R}^{d\times K_l}$ as the cluster centroids and $\mathbf{H}_l \in \mathbf{R}^{K_l \times n}$ as the cluster indicators, or separately clustering on $\mathbf{H}_l$, we can obtain $m$ clusterings by a deep factorization network with $m$ layers. However, these clusterings may have high redundancy, since the overlap between them is ignored.

Multi-view data often embody different distributions, which enable different groupings of the same dataset from diverse perspectives. Therefore, it is promising to apply DMF on multi-view data to discover multiple clusterings. One simple solution is to concatenate multiple feature views into a single view, and then directly apply DMF on the concatenated view. However, this concatenation does not differentiate the relevance of these views, and results in information override and redundant clusterings.  Given that, we propose the multi-view multiple clusterings using deep matrix factorization solution.

\subsection{The proposed method}
\label{sec:mvmc}
Suppose $\mathcal{X}=\{\mathbf{X}^{(1)}, \mathbf{X}^{(2)}, \cdots, \mathbf{X}^{(V)}\}$ is a dataset with $V$ different feature views of $n$ objects, $\mathbf{X}^{(v)} \in \mathbb{R}^{d_v \times n}$. To make use of the complementary information and to explore hierarchical representations of multi-view data, we formulate our model by extending DMF as follows:
\begin{equation}
\small
 \begin{split}
\underset{\mathbf{Z}_m^{(v)}, \mathbf{H}_m}{min}&=\sum_{m=1, m'\neq m}^M \sum_{v=1}^V||\mathbf{X}^{(v)}-\mathbf{Z}_1^{(v)}\mathbf{Z}_2^{(v)}...\mathbf{Z}_m^{(v)}\mathbf{H}_m||_F^2\\
&+ \lambda \mathcal{R}(\mathbf{H}_m,\mathbf{H}_{m'})
\end{split}
\label{Eq2}
\end{equation}
where $M$ is the user-specified target number of clusterings, $\mathbf{Z}_l^{(v)}$ is the $l$-th ($l \leq m$) layer mapping for view $v$, $\mathcal{R}(\mathbf{H}_{m},\mathbf{H}_{m'})$ quantifies the redundancy between two clusterings and will be discussed later. $\lambda$ is introduced to balance the quality and redundancy of $M$ clusterings. Since $\mathbf{H}_m$ is shared across all the data views, we can expect that $\mathbf{H}_m$ fuses the complementary information of multiple data views to generate a high-quality representational subspace in the $m$-th layer with respect to $\mathbf{Z}_1^{(v)}\mathbf{Z}_2^{(v)}...\mathbf{Z}_m^{(v)}$. In addition, because of the hierarchical representation and redundancy control term, alternative clusterings with diversity can be pursued also.

Our formulation has a close connection with multi-view clustering via deep matrix factorization \cite{zhao2017dmf}, which also factorizes multiple data views layer-by-layer to extract the complementary information, but it can only generate a single clustering in the final layer. Our task is different from subspace clustering \cite{domeniconi2007locally,luo2018consistent}, which seeks only one clustering with different clusters in different subspaces. Our formulation is also different from non-redundant multiple clustering by nonnegative matrix factorization (MNMF) \cite{yang2017MNMF}, which performs only one layer factorization to find a new clustering by reducing the redundancy between the clustering and already explored ones. As such, MNMF may generate low quality alternative clusterings due to its one-layer representation of data and the heavy dependence on the reference clustering.

Different data views may have a different relevance toward different clusterings. Eq. (\ref{Eq2}) and MVMC \cite{yao2019MVMC} assume all the data views have the same relevance toward these clusterings. As such, the noisy or irrelevant data views may compromise the quality of alternative clusterings. To account for the different levels of relevance of the data views toward the alternative clusterings, and reduce the impact of noisy views, we further assign weights to these views for each clustering as follows:
\begin{equation}
\small
\begin{split}
\underset{\mathbf{Z}_l^{(v)}, \mathbf{H}_m,  \alpha_m^{(v)}}{min} &=\sum_{m=1, m'>m}^M \sum_{v=1}^V (\alpha_m^{(v)})^r ||\mathbf{X}^{(v)}-\mathbf{Z}_m^{(v)}...\mathbf{Z}_m^{(v)}\mathbf{H}_m||_F^2\\
&+ \lambda \mathcal{R}(\mathbf{H}_{m},\mathbf{H}_{m'})\\
& s.t., \ \mathbf{H}_m \geq 0, \ {\sum}_{v=1}^V \alpha^{(v)}_m=1, \ \alpha^{(v)}_m\geq 0
\end{split}
\label{Eq3}
\end{equation}
where $\alpha_m^{(v)} \geq 0$ is the weight coefficient for the $v$-th data view for generating the $m$-th clustering, and $r$ is the parameter to control the weights distribution. In this way, multiple data views are selectively fused to generate diverse clustering with quality. {For example, in Figure \ref{fig:case}, three alternative clusterings (shape, color, texture) can be obtained by different weight assignments of three views.}

As we stated, it is important to control the redundancy (or overlap) with alternative clusterings.
Most subspace based multiple clusterings solutions reduce the redundancy between clusterings by seeking orthogonal (non-redundant or independent) subspaces \cite{cui2007OSC,ye2016generalized,mautz2018NrKMeans,wang2019MISC}. DMClusts also has such flavor and seeks a clustering based on each layer's representation $\mathbf{H}_m$.  However, a set of objects maybe nearby in the orthogonally projected subspaces and thus outputs similar clusters in these subspaces. For this reason, we additionally quantify the redundancy between clusterings using ($\{\mathbf{H}_m\}_{m=1}^M$). A co-association matrix $\mathbf{C}^{(m)} \in \mathbb{R}^{n \times n}$ can reflect whether two objects are grouped into the same cluster or not for the $m$-th clustering \cite{fred2005coassociation}. Particularly, if $\mathbf{x}_i$ and $\mathbf{x}_j$ are grouped into the same cluster, then $\mathbf{C}^{(m)}_{ij}=1$, otherwise $\mathbf{C}^{(m)}_{ij}=0$. So if two clusterings ($m$ and $m'$) have a large ${\sum}_{ij}^n \mathbf{C}^m_{ij}  \mathbf{C}^{m'}_{ij}$, there is a high redundancy (or overlap) between them. Since the normalized representation $\mathbf{H}_m$ often can not be an exact binary cluster-indicator matrix, here we approximate $\mathbf{C}^{(m)}$ by $\mathbf{H}_m^T\mathbf{H}_m$, which softly quantifies the degree of two objects being grouped into the same cluster for the $m$-th layer (or clustering). Based on this approximation, we quantify the overlap between two clusterings in different layers as:
\begin{equation}
\small
\centering
\begin{split}
\mathcal{R}(\mathbf{H}_{m},\mathbf{H}_{m'})&={\sum}_{i,j=1}^n(\mathbf{H}_m^T \mathbf{H}_m)_{ij}(\mathbf{H}_{m'}^T \mathbf{H}_{m'})_{ij}\\
&=tr(\mathbf{H}_m^T \mathbf{H}_m\mathbf{H}_{m'}^T \mathbf{H}_{m'})
\end{split}
\label{Eq4}
\end{equation}
where $tr(\cdot)$ is the matrix trace operator. A large  $\mathcal{R}(\mathbf{H}_{m},\mathbf{H}_{m'})$ means $\mathbf{x}_i$ and $\mathbf{x}_j$ are nearby in different representation subspaces, which will be grouped into the same clusters of two different clusterings and increase the overlap.

However, Eq. (\ref{Eq4}) \emph{only} accounts for the case that two objects are often projected nearby (grouped into the same clusters) in different representation subspaces, but overlooks the case that two objects are frequently placed faraway (grouped into different clusters) in these subspaces. We want to remark that other multiple clustering solutions \cite{yang2017MNMF,wang2018mcc,yao2019mccss} also adopt the idea in Eq. (\ref{Eq4}) to quantify the redundancy between clusterings, and thus they also overlook the latter case, which emerges when the number of clusters $\geq 3$. To remedy this overlook, we introduce a \emph{balanced} redundancy quantification term as follows:
\begin{eqnarray}
\small
\centering
\begin{split}
\tilde{\mathcal{R}}(\mathbf{H}_{m},\mathbf{H}_{m'})&=\beta tr(\mathbf{H}_m^T \mathbf{H}_m \mathbf{H}_{m'}^T\mathbf{H}_{m'})\\
&+(1-\beta)tr((1-\mathbf{H}_m^T \mathbf{H}_m)(1-\mathbf{H}_{m'}^T\mathbf{H}_{m'})\\
\end{split}
\label{Eq5}
\end{eqnarray}
where $\beta \in [0,1]$ is the balance coefficient. Eq. (\ref{Eq5}) considers two extreme cases: (i) many pairwise objects are always \emph{nearby} in two subspaces, (ii) are always \emph{faraway} in these subspaces. Both cases increase the overlap of two clusterings. In other words, if many pairwise objects placed into the same clusters for one clustering, but not so for the other clustering, then the redundancy between them is low.

To this end, we can reformulate the objective function of DMClusts as follows:
\begin{equation}
\small
\centering
\begin{split}
\min\limits_{\mathbf{Z}_m^{(v)}, \mathbf{H}_m,  \alpha_m^{(v)}}\mathcal{J}&= \sum_{m=1, m\neq m'}^M \sum_{v=1}^V (\alpha_m^{(v)})^r ||\mathbf{X}^{(v)}-\mathbf{Z}_1^{(v)}...\mathbf{Z}_m^{(v)}\mathbf{H}_m||_F^2\\
&+ \lambda (\beta tr(\mathbf{H}_m^T \mathbf{H}_m \mathbf{H}_{m'}^T\mathbf{H}_{m'}) \\
&+ (1-\beta)tr((1-\mathbf{H}_m^T \mathbf{H}_m)^T(1-\mathbf{H}_{m'}^T\mathbf{H}_{m'})))\\
& s.t., \ \mathbf{H}_m \geq 0, \ {\sum}_{v=1}^V \alpha^{(v)}_m=1, \ \alpha^{(v)}_m\geq 0
\end{split}
\label{Eq6}
\end{equation}
By minimizing the above objective, we can gradually find $M$ clusterings, while the quality of these clusterings is pursued by the constraint $\mathbf{H}_m$ of the respective layer shared across all the views, and the diversity is pursued by reducing the cases that too many objects always nearby (or faraway) in these representation subspaces. Our experiments will confirm the advantage of these factors.

\subsection{Optimization}
\label{sec:mvmcopt}
The minimization objective in Eq. (\ref{Eq6}) is defined with respect
to $\mathbf{Z}_m^{(v)}$, $\mathbf{H}_m$, and $\alpha_m^{(v)}$. Since a close-form solution cannot be
given, we alternatively optimize one variable while keeping the other two constant. The alternative process is detailed below.

\textbf{Update rule for $\mathbf{Z}_m^{(v)}$}: The optimization of Eq. (\ref{Eq6}) with respect to $\mathbf{Z}_m^{(v)}$ is:
\begin{equation}
\small
\centering
\begin{split}
\mathcal{J}_Z(\mathbf{Z}_{m}^{(v)})&=\sum_{i=m}^M(\alpha_i^{(v)})^r||\mathbf{X}^{(v)}-\Phi_{m}\mathbf{Z}_{m}^{(v)}\mathbf{H}_{mi}||_F^2\\
\end{split}
\label{Eq7}
\end{equation}
where $\Phi_{m}=\mathbf{Z}_{1}^{(v)}\mathbf{Z}_{2}^{(v)}...\mathbf{Z}_{m-1}^{(v)}$ and $\mathbf{H}_{mi}=\mathbf{Z}_{m+1}^{(v)}...\mathbf{Z}_{i}^{(v)}\mathbf{H}_{i}$. Letting the partial derivative $\partial\mathcal{J}_Z/\mathbf{Z}_{m}^{(v)}$=0, we can obtain
\begin{equation}
\small
\begin{split}
\mathbf{Z}_{m}^{(v)}&=(\Phi_m^{T}\Phi_m)^{-1}(\sum_{i=m}^M(\alpha_i^{(v)})^r{\Phi}^{T}\mathbf{X}^{(v)}\mathbf{H}_{mi}^{T})\\
&(\sum_{i=m}^M(\alpha_i^{(v)})^r\mathbf{H}_{mi}\mathbf{H}_{mi}^{T})^{-1}\\
\end{split}
\label{Eq8}
\end{equation}

\textbf{Update rule for $\mathbf{H}_m$}: Optimizing Eq. (\ref{Eq6}) with respect to $\mathbf{H}_{m}$ is equivalent to minimizing the following:
\begin{equation}
\small
\begin{split}
\mathcal{J}_H(\mathbf{H}_m)&=\sum_{v=1}^V \alpha_m^{(v)}||\mathbf{X}^{(v)}-\mathbf{Z}_{1}^{(v)}...\mathbf{Z}_{m}^{(v)}\mathbf{H}_{m}||_F^2\\
&+ \lambda (\beta tr(\mathbf{H}_m^T \mathbf{H}_m \mathbf{H}_{m'}^T\mathbf{H}_{m'}) \\
&+ (1-\beta)tr((1-\mathbf{H}_m^T \mathbf{H}_m)^T(1-\mathbf{H}_{m'}^T\mathbf{H}_{m'})))\\
& s.t., \ \mathbf{H}_m \geq 0
\end{split}
\label{Eq9}
\end{equation}
For the constraint $\mathbf{H}_m\ge0$, we introduce the Lagrangian multiplier $\eta$ as follows:
\begin{equation}
\small
\centering
\begin{split}
\mathcal{L}(\mathbf{H}_m)&=\sum_{v=1}^V \alpha_m^{(v)} ||\mathbf{X}^{(v)}-\mathbf{Z}_{1}^{(v)}...\mathbf{Z}_{m}^{(v)}\mathbf{H}_m||_F^2\\
&+\lambda \sum_{m'=1,m'\neq m}^M \beta tr(\mathbf{H}_m^T\mathbf{H}_m\mathbf{H}_{m'}^T\mathbf{H}_{m'} )\\
&+(1-\beta)tr((1-\mathbf{H}_m^T\mathbf{H}_m)(1-\mathbf{H}_{m'}^T \mathbf{H}_{m'}))-tr(\eta \mathbf{H}_m)\\
\end{split}
\label{Eq10}
\end{equation}

Letting the partial derivative $\partial \mathcal{L}/{\mathbf{H}_m}=0$ and $\eta_{ij}(\mathbf{H}_m)_{ij}=0$, we can get
\begin{equation}
\small
\centering
\begin{split}
\mathbf{H}_{m}&=\mathbf{H}_m \odot \sqrt{\frac{\mathbf{Q}^{+}+\mathbf{P}^{-}\mathbf{H}_m+\lambda\Gamma_m^-}
{\mathbf{Q}^{-}+\mathbf{P}^{+}\mathbf{H}_m+\lambda\Gamma_m^+}}
\end{split}
\label{Eq11}
\end{equation}
where  $\mathbf{Q}={\sum}_{v=1}^V(\alpha_m^{(v)})^r (\mathbf{Z}_{all}^{(v)})^T \mathbf{X}^{(v)}$, $\mathbf{P}={\sum}_{v=1}^V(\alpha_m^{(v)})^r (\mathbf{Z}_{all}^{(v)})^T \mathbf{Z}_{all}^{(v)}$. $\mathbf{Q}_{ij}^{+}=({|\mathbf{Q}|}_{ij}+\mathbf{Q}_{ij})/2$, $\mathbf{Q}_{ij}^{-}=({|\mathbf{Q}|}_{ij}-\mathbf{Q}_{ij})/2$, $\Gamma_m={\sum}_{m=1,m'\neq m}^M \mathbf{H}_{m}\mathbf{H}_{m'}^T\mathbf{H}_{m'}-(1-\beta)\mathbf{H}_{m}\mathbf{1}^{T}$, $\mathbf{Z}_{all}^{(v)}=\mathbf{Z}_1^{(v)} \mathbf{Z}_2^{(v)}...\mathbf{Z}_{m}^{(v)}$.

\textbf{Update rule for $\alpha_m^{(v)}$}: We denote ${\Theta}_m^{(v)}=||\mathbf{X}^{(v)}-\mathbf{Z}_{1}^{(v)}...\mathbf{Z}_{m}^{(v)}\mathbf{H}_{m}||_F^2$. Eq. (\ref{Eq6}) with respect to $\alpha_m^{(v)}$ is written as:
\begin{equation}
\small
\centering
\begin{split}
\mathop{min}_{\alpha_m^{(v)}}\sum_{v=1}^V(\alpha_m^{(v)})^r {\Theta}_m^{(v)} \ \textrm{s.t.} \sum_{v=1}^V \alpha_m^{(v)}=1, \alpha_m^{(v)}\ge0. \\
\end{split}
\label{Eq12}
\end{equation}
The Lagrangian function of Eq. (\ref{Eq12}) is:
\begin{equation}
\small
\centering
\begin{split}
\mathop{min}_{\alpha_m^{(v)}}\sum_{v=1}^V(\alpha_m^{(v)})^r {\Theta}_m^{(v)}-\lambda(\sum_{v=1}^V\alpha_m^{(v)}-1). \\
\end{split}
\label{Eq13}
\end{equation}
where $\lambda$ is the Lagrangian multiplier. By taking the derivative of Eq. (\ref{Eq13}) with respect to $\alpha_m^{(v)}$, and setting it to zero, we have $\alpha_m^{(v)}=({\lambda}/{r\Theta_m^{(v)}})^{\frac{1}{r-1}}$. Since $\sum_{v=1}^V(\alpha_m^{(v)})^r=1$, we can obtain:
\begin{equation}
\small
\alpha_m^{(v)}={(r \Theta_m^{(v)})^{\frac{1}{1-r}}}/{{\sum}_{v=1}^V(r \Theta_m^{(v)})^{\frac{1}{1-r}}}
\label{Eq15}
\end{equation}
To this end, we have all the iterative update rules for optimizing three variables of DMClusts. We repeat these updates iteratively until convergence. After that, we run $k$-means clustering on each $\{\mathbf{H}_m\}_{m=1}^M$ and obtain $M$ clusterings.

\subsection{Time complexity}
The time complexity of DMClusts is composed of three parts. For simplicity, we assume all the layers have the same  size $K$. DMClusts takes order $\mathcal{O}(M(ndK+dK^2+nK^2))$ to update $\mathbf{Z}_m^{(v)}$,   $\mathcal{O}(V ndK)$ to update $\alpha_m^{(v)}$, and $\mathcal{O}(V ndK+MnK^2)$ to update $\mathbf{H}_m$ in each iteration. So the time complexity of DMClusts for generating $M$ clusterings on $V$ views is  $\mathcal{O}(tM(MndK+MnK^2+MdK^2+VndK))$, where $t$ is the number of iterations to convergence.  Generally $K<d$, $K<n$, and $M\ll n$, thus the complexity of DMClusts is $\mathcal{O}(t(M^2+V)ndK)$. {In our used datasets, DMClusts converges within $t<50$ iterations.} On the other hand, the time complexity of MVMC \cite{yao2019MVMC} is $\mathcal{O}(tVM(n^2d+n^2k))$ ($k$ is the number of clusters). Clearly, the complexity of DMClusts is linear in $n$, but MVMC is quadratic to $n$. As a result, our DMClusts can scale to larger datasets than MVMC.

\section{Experimental Results and Analysis}
\subsection{Experimental Setup}
In this section, we evaluate the effectiveness and efficiency of our proposed DMClusts on seven widely-used multi-view datasets, as described in  Table \ref{table1}. The adopted datasets are from different domains, with different numbers of views and objects. More details on the data are given in the Supplementary file.

Multiple clustering approaches aim to achieve diverse clusterings of high quality.
To measure quality, we use Silhouette Coefficient (SC) and the Dunn Index (DI) as  internal indexes to quantify the  compactness and separation of clusters.
To measure redundancy, we use Normalized Mutual Information (NMI) and Jaccard Coefficient (JC) as  external indexes to quantify  the similarity of clusters between two clusterings. We want to emphasize that a \emph{higher} value of SC and DI means a clustering with  \emph{higher} quality, but a \emph{smaller} value of NMI and JC implies that two clusterings have a \emph{smaller} redundancy. These metrics have been widely adopted  for evaluating multiple clusterings \cite{bailey2013alternative,yang2017MNMF}. Their formal definitions are given in the Supplementary file.
\begin{table}[h!tbp]
\scriptsize
		\caption{Statistics of multi-view datasets. $n$, $c$, $V$ are the numbers of objects, clusters and views; $d_v$ are the dimensions of $V$ views.}
		\centering
\begin{tabular}{l|l |l}
	\hline
	Datasets &$n$, $c$, $V$ &$d_v$\\
	\hline
    Caltech7 &1474, 7, 6 &[40, 48, 254, 1984, 512, 928]\\
    Handwritten &2000, 10, 6 &[216, 76, 64, 6, 240, 47]\\
    Reuters &1200, 6, 5 & [21531, 24892, 34251, 15506, 11547]\\
    BBCSport &145, 2, 4 &[4659, 4633, 4665, 4684]\\
    MSRCv1 &210, 7, 6 &[1302, 48, 512, 100, 256, 210]\\
    Yale &165, 15, 3 &[4096, 3304, 6750]\\
    Mirflickr &16738, 24, 2 &[150, 500]\\

	\hline
\end{tabular}
\label{table1}
\end{table}
\subsection{Discovering multiple clusterings}
To comparatively study the performance of DMClusts, we consider Dec-kmeans \cite{jain2008deckmeans}, MVMC \cite{yao2019MVMC}, OSC \cite{cui2007OSC},  ISAAC \cite{ye2016generalized}, MNMF \cite{yang2017MNMF}, and MISC \cite{wang2019MISC} as comparing methods. The last four  methods use different techniques to seek clusterings in subspaces. The input parameters of the comparing methods are fixed (or optimized) as the authors suggested in their papers or shared code.
{The input parameters of DMClusts are selected from the following ranges: $r \in \{ 5\times 10^{-4}, 5\times 10^{-3}, ..., 5\}$, $\lambda \in \{10^{-4}, 10^{-3},..., 10^{4}\}$, $\beta \in [0, 1]$ and $K_1 \in [k, min(d_v)]$, $K_2 \in [k, K_1]$ with $M=2$.}  We fix the number of clusters for each clustering to the number of classes $c$ of each dataset, as reported in Table \ref{table1}. Existing multiple clustering algorithms (except MVMC and DMClusts) cannot work on multiple view data. Following the solution in \cite{yao2019MVMC}, we concatenate the feature vectors of multi-view data and then run them on the concatenated vectors to seek alternative clusterings. For reference, we also apply DMF \cite{trigeorgis2017deep} on the concatenated vectors to gradually explore multiple clusterings layer by layer.

MNMF requires input a reference clustering to find an alternative clustering. Here we use $k$-means to generate the reference clustering. For the other comparing methods, we directly use their respective solutions to generate two alternative clusterings ($\mathcal{C}_1$, $\mathcal{C}_2$).
Following the evaluation protocol used by the comparing methods, we measure  clustering quality  with the average (SC or DI) of $\mathcal{C}_1$ and $\mathcal{C}_2$, and we measure the diversity (NMI or JC) between $\mathcal{C}_1$ and $\mathcal{C}_2$.  Table\ref{table2} gives the average results of ten independent runs and standard deviations of each method on generating two alternative clusterings. The results of ISAAC and MISC on Reuters and Mirflickr are not reported for their high complexity on large scale datasets.

From Table\ref{table2}, we make the following observations:\\
(i) \textbf{Multi-view vs. Concatenated view}: Both DMClusts and MVMC directly operate on multi-view data, and their generated two clusterings have a significant lower redundancy than those generated by other comparing methods. In addition, DMClusts frequently obtains a better quality than other comparing methods that can only work on the concatenated view. This shows that the concatenated feature vectors override the intrinsic nature of multi-view data, which help to generate multiple clusterings with diversity. This also expresses the capability of our tailored deep matrix factorization in exploring multiple clusterings with quality.\\
(ii) \textbf{DMClusts vs. MVMC}: DMClusts generally obtains a significantly better quality (SC and DI) than MVMC, and holds a comparable diversity (NMI and JC). In other words, our DMClusts maintains a better balance of quality and diversity than MVMC. A possible factor is that DMClusts differentiates the relevance of multiple views, whereas MVMC does not. As a result, DMClusts is less sensitive to the noisy views than MVMC. Another factor is that our balanced redundancy term is more comprehensive by considering two types of redundancy, but MVMC considers only one type.\\
(iii) \textbf{DMClusts vs. DMF}: DMClusts always gives a better performance  (both quality and diversity) than DMF, although they both can explore alternative clusterings in a layer-wise fashion. The advantage of DMClusts is two-fold: it accounts for the different relevance of data views, and can selectively fuse them to generate alternative clusterings with quality, while DMF can only operate on the concatenated features without differentiating these views;  it also explicitly controls the diversity between alternative clusterings, while DMF does not.

\begin{figure}[h!tbp]
\centering
\includegraphics[width=0.66\columnwidth]{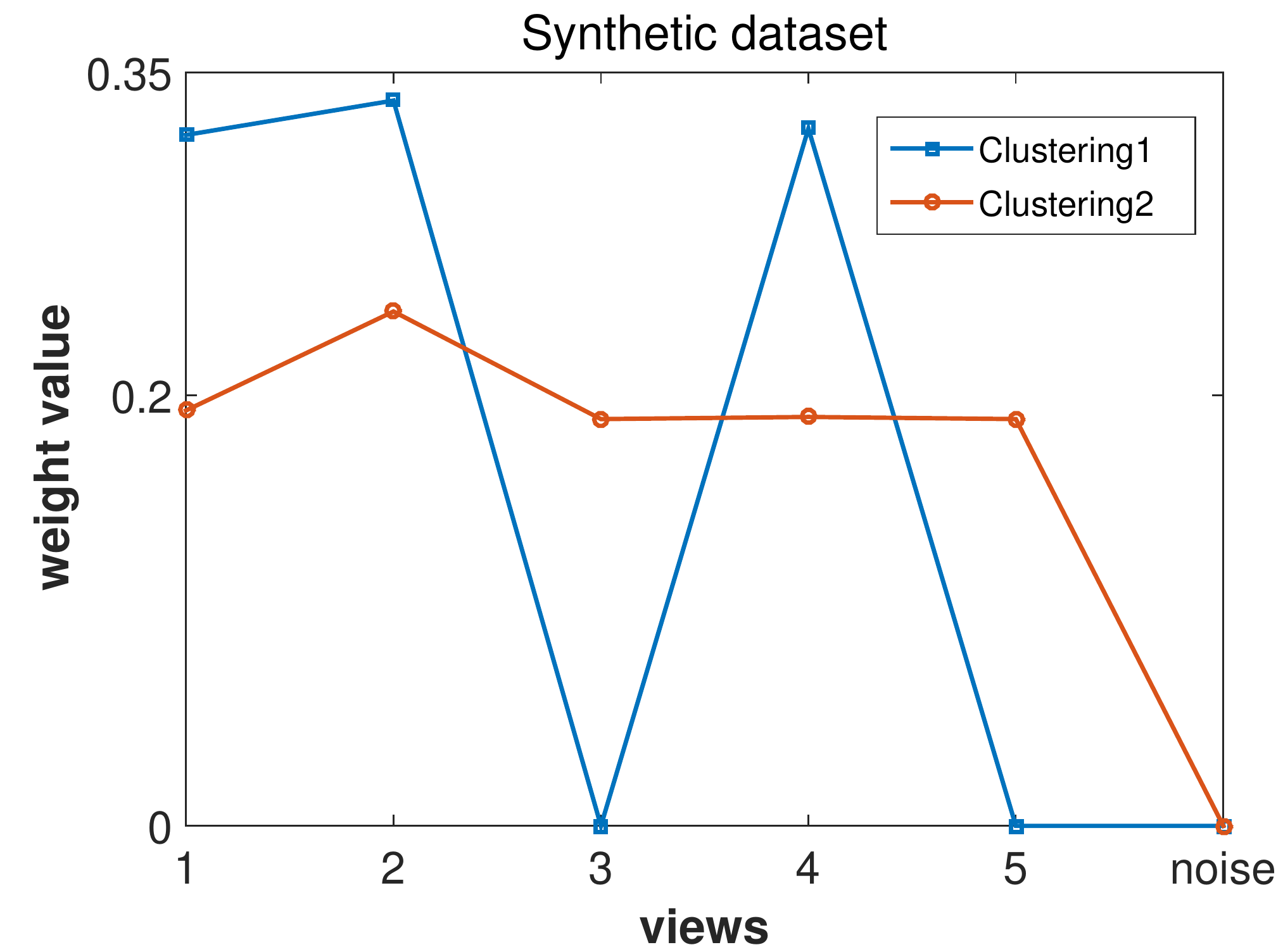}
\caption{DMClusts assigns two sets of weights to six views for generating two clusterings. The 6-th view is a \emph{noisy} view.}
\label{figNoisy}
\end{figure}

\begin{table*}[t]
\scriptsize
	\caption{Quality and Diversity of the various comparing methods on generating multiple clusterings. $\uparrow$($\downarrow$) indicates the preferred direction for the corresponding measure. $\bullet / \circ$ indicates whether our DMClusts is statistically (according to pairwise $t$-test at 95\% significance level) superior/inferior to the other method.}
    \centering
	\resizebox{1.8\columnwidth}{!}{
	\begin{tabular}{c|c| r r r r r r r |r}
		\hline
		& &Dec-kmeans &MNMF &OSC &ISAAC &MISC &MVMC &DMF &DMClusts\\
		\hline
        \multirow{4}[2]{*}{Caltech7}
        &SC$\uparrow$
        & 0.049$\pm$0.002$\bullet$ &0.234$\pm$0.000$\bullet$ & 0.266$\pm$0.000$\bullet$ & 0.153$\pm$0.010$\bullet$ & 0.201$\pm$0.003$\bullet$ & 0.140$\pm$0.002$\bullet$ &0.065$\pm$0.011$\bullet$
        & 0.301$\pm$0.006 \\
        &DI$\uparrow$
        & 0.042$\pm$0.006$\bullet$ & 0.037$\pm$0.000$\bullet$ & 0.054$\pm$0.000$\bullet$ & 0.027$\pm$0.001$\bullet$ & 0.048$\pm$0.002$\bullet$ & 0.062$\pm$0.000$\bullet$ &0.065$\pm$0.004$\bullet$ &0.090$\pm$0.003 \\
        &NMI$\downarrow$
        & 0.021$\pm$0.003$\bullet$ & 0.022$\pm$0.000$\bullet$ & 0.693$\pm$0.015$\bullet$ & 0.645$\pm$0.035$\bullet$ & 0.516$\pm$0.015$\bullet$ & 0.006$\pm$0.000$\circ$ &0.310$\pm$0.035$\bullet$ &0.009$\pm$0.000 \\
        &JC$\downarrow$
        & 0.127$\pm$0.009$\bullet$ & 0.092$\pm$0.000$\bullet$ & 0.383$\pm$0.000$\bullet$ & 0.358$\pm$0.022$\bullet$ & 0.349$\pm$0.018$\bullet$ & 0.076$\pm$0.000$\circ$ &0.235$\pm$0.004$\bullet$ &0.087$\pm$0.002 \\
        \hline
        \multirow{4}[2]{*}{BBCSport}
        &SC$\uparrow$
        &0.088$\pm$0.007$\bullet$ &0.014$\pm$0.000$\bullet$ &0.144$\pm$0.000$\bullet$ &-0.039$\pm$0.002$\bullet$ &0.089$\pm$0.002$\bullet$ &0.269$\pm$0.000$\bullet$ &0.204$\pm$0.003$\bullet$ &0.284$\pm$0.006 \\
        &DI$\uparrow$
        &0.487$\pm$0.001$\circ$ &0.434$\pm$0.000$\bullet$ &0.520$\pm$0.000$\circ$ &0.411$\pm$0.016$\bullet$ &0.335$\pm$0.009$\bullet$ &0.014$\pm$0.000$\bullet$ &0.255$\pm$0.001$\bullet$ &0.468$\pm$0.010 \\
        &NMI$\downarrow$
        &0.002$\pm$0.000$\bullet$ &0.086$\pm$0.000$\bullet$ &0.001$\pm$0.000$\bullet$ &0.010$\pm$0.001$\bullet$ &0.009$\pm$0.001$\bullet$ &0.000$\pm$0.000
        &0.101$\pm$0.009 $\bullet$
        &0.000$\pm$0.000 \\
        &JC$\downarrow$
        &0.431$\pm$0.030$\bullet$ &0.392$\pm$0.000$\circ$ &0.605$\pm$0.000$\bullet$ &0.495$\pm$0.015$\bullet$ &0.520$\pm$0.018$\bullet$ &0.347$\pm$0.000$\circ$ &0.418$\pm$0.001$\bullet$ &0.401$\pm$0.003 \\
        \hline
        \multirow{4}[2]{*}{Handwritten}
        &SC$\uparrow$
        &0.050$\pm$0.006$\bullet$ &0.014$\pm$0.000$\bullet$ &0.352$\pm$0.000$\bullet$ &0.235$\pm$0.007$\bullet$ &0.251$\pm$0.009$\bullet$ &0.062 $\pm$0.000$\bullet$ &0.034$\pm$0.001$\bullet$ &0.377$\pm$0.012 \\
        &DI$\uparrow$
        &0.051$\pm$0.011$\bullet$ &0.009$\pm$0.000$\bullet$ &0.107$\pm$0.000$\bullet$ &0.056$\pm$0.002$\bullet$ &0.052$\pm$0.003$\bullet$ &0.083$\pm$0.000$\bullet$ &0.240$\pm$0.009$\circ$ &0.159$\pm$0.004\\
        &NMI$\downarrow$
        &0.070$\pm$0.012$\bullet$ &0.089$\pm$0.000$\bullet$ &0.778$\pm$0.000$\bullet$ &0.712$\pm$0.018$\bullet$ &0.645$\pm$0.014$\bullet$ &0.009$\pm$0.000$\circ$ &0.212$\pm$0.006$\bullet$ &0.019$\pm$0.001 \\
        &JC$\downarrow$
        & 0.073$\pm$0.003$\bullet$ &0.078$\pm$0.003$\bullet$ &0.570$\pm$0.000$\bullet$ &0.484$\pm$0.016$\bullet$ &0.414$\pm$0.019$\bullet$ &0.073$\pm$0.000$\bullet$ &0.114$\pm$0.003$\bullet$ &0.066$\pm$0.000 \\
        \hline
        \multirow{4}[2]{*}{MSRCv1}
        &SC$\uparrow$
        &-0.062$\pm$0.003$\bullet$ &-0.193$\pm$0.002$\bullet$ &0.382$\pm$0.011$\bullet$ &0.166$\pm$0.003$\bullet$ &0.331$\pm$0.008$\bullet$ &0.113$\pm$0.007$\bullet$ &0.022$\pm$0.001$\bullet$ &0.556$\pm$0.012 \\
        &DI$\uparrow$
        &0.043$\pm$0.007$\bullet$ &0.027$\pm$0.001$\bullet$ &0.071$\pm$0.007$\bullet$
        &0.012$\pm$0.002$\bullet$ &0.013$\pm$0.001$\bullet$ &0.098$\pm$0.003$\bullet$ &0.277$\pm$0.010$\bullet$ &0.336$\pm$0.008 \\
        &NMI$\downarrow$
        &0.054$\pm$0.006$\bullet$ &0.063$\pm$0.005$\bullet$ &0.736$\pm$0.054$\bullet$ &0.549$\pm$0.030$\bullet$ &0.665$\pm$0.017$\bullet$ &0.053$\pm$0.006$\bullet$ &0.150$\pm$0.002$\bullet$ &0.038$\pm$0.001 \\
        &JC$\downarrow$
        &0.109$\pm$0.005$\bullet$ &0.124$\pm$0.003$\bullet$ &0.519$\pm$0.025$\bullet$ &0.357$\pm$0.009$\bullet$ &0.471$\pm$0.020$\bullet$ &0.078$\pm$0.002$\circ$ &0.127$\pm$0.005$\bullet$ &0.087$\pm$0.003 \\
        \hline
        \multirow{4}[2]{*}{Yale}
        &SC$\uparrow$
        &0.033$\pm$0.002$\bullet$ &-0.011$\pm$0.001$\bullet$ &0.221$\pm$0.005$\bullet$ &-0.020$\pm$0.002$\bullet$ &-0.066$\pm$0.008$\bullet$ &-0.045$\pm$0.007$\bullet$ &0.021$\pm$0.001$\bullet$ &0.303$\pm$0.019 \\
        &DI$\uparrow$
        &0.205$\pm$0.014$\bullet$ &0.114$\pm$0.004$\bullet$ &0.331$\pm$0.020$\circ$ &0.076$\pm$0.004$\bullet$ &0.073$\pm$0.003$\bullet$ &0.232$\pm$0.012$\bullet$ &0.285$\pm$0.004$\bullet$ &0.292$\pm$0.015 \\
        &NMI$\downarrow$
        &0.241$\pm$0.021$\bullet$ &0.240$\pm$0.007$\bullet$ &0.812$\pm$0.063$\bullet$ &0.369$\pm$0.007$\bullet$ &0.314$\pm$0.009$\bullet$ &0.251$\pm$0.006$\bullet$ &0.319$\pm$0.006$\bullet$ &0.205$\pm$0.004 \\
        &JC$\downarrow$
        &0.043$\pm$0.002$\bullet$ &0.066$\pm$0.004$\bullet$ &0.357$\pm$0.034$\bullet$ &0.098$\pm$0.003$\bullet$ &0.091$\pm$0.002$\bullet$
        & 0.055$\pm$0.001$\bullet$ &0.098$\pm$0.005$\bullet$ &0.038$\pm$0.002 \\
        \hline
        \multirow{4}[2]{*}{Reuters}
        &SC$\uparrow$
        & -0.002$\pm$0.000$\bullet$ &-0.107$\pm$0.009$\bullet$ &0.065$\pm$0.000$\bullet$ &--- &--- &0.180$\pm$0.000$\bullet$ &0.314$\pm$0.004$\bullet$ &0.344$\pm$0.006 \\
        &DI$\uparrow$
        &0.157$\pm$0.008$\circ$ &0.070$\pm$0.003$\bullet$ &0.210$\pm$0.000$\circ$ &--- &--- &0.038$\pm$0.000$\bullet$ &0.028$\pm$0.001$\bullet$ &0.136$\pm$0.005\\
        &NMI$\downarrow$
        &0.041$\pm$0.004$\bullet$ &0.033$\pm$0.010$\bullet$ &0.491$\pm$0.000$\bullet$ &--- &--- &0.004$\pm$0.000$\circ$ &0.508$\pm$0.005$\bullet$ &0.018$\pm$0.000 \\
        &JC$\downarrow$
        &0.199$\pm$0.005$\bullet$ &0.148$\pm$0.002$\bullet$ &0.454$\pm$0.000$\bullet$ &--- &--- &0.091$\pm$0.000$\circ$ &0.590$\pm$0.011$\bullet$ &0.132$\pm$0.003 \\
        \hline
        \multirow{4}[2]{*}{Mirflicker}
        &SC$\uparrow$
        & -0.004$\pm$0.000$\bullet$ &-0.058$\pm$0.000$\bullet$ &0.017$\pm$0.000$\bullet$ &--- &--- &-0.038$\pm$0.000$\bullet$ &0.005$\pm$0.000$\bullet$
        &0.336$\pm$0.008\\
        &DI$\uparrow$
        &0.061$\pm$0.002$\bullet$ &0.053$\pm$0.001$\bullet$ &0.059$\pm$0.002$\bullet$ &--- &--- &0.173$\pm$0.005$\circ$ &0.027$\pm$0.001$\bullet$ &0.076$\pm$0.001\\
        &NMI$\downarrow$
        &0.427$\pm$0.012$\bullet$ &0.014$\pm$0.000$\bullet$ &0.575$\pm$0.011$\bullet$ &--- &--- &0.005$\pm$0.000$\circ$ &0.108$\pm$0.003$\bullet$ &0.043$\pm$0.001 \\
        &JC$\downarrow$
        &0.878$\pm$0.022$\bullet$ &0.023$\pm$0.000$\circ$ &0.368$\pm$0.011$\bullet$ &--- &--- &0.022$\pm$0.000$\circ$ &0.049$\pm$0.001$\bullet$ &0.033$\pm$0.001 \\
        \hline
\end{tabular}}
\label{table2}
\end{table*}
To investigate the robustness of DMClusts to noisy views, {we constructed a synthetic dataset on Reuters by injecting a noisy view $\mathbf{X}^{(6)} \in \mathbb{R}^{500 \times 1200}$ following standard Gaussian distribution. We then apply DMF and DMClusts on this synthetic dataset with the input parameters fixed as $r=0.1$, $\lambda=0.1$, $\beta=0.4$}. Next, we visualize the weights assigned to six views for the first and second clusterings in Figure \ref{figNoisy}. DMClusts indeed assigns different sets of weights to these views for generating two clusterings with a low overlap (NMI: 0.019, JC: 0.161), and it manifests a robustness to the noisy view by assigning it with a zero weight. As a result, DMClusts holds the similar  quality and diversity as on the original Reuters. In contrast, DMF has a nearly 50\% reduced quality (SC: 0.158, DI: 0.015) and an about 25\% increased diversity (NMI: 0.471, JC: 0.375). The increase in diversity is obtained at the expense of a reduced quality. Nevertheless, DMClusts still gives a better diversity than DMF. This investigation corroborates the benefit of weighting views.

\begin{figure*}[t]	
	\centering
	\begin{subfigure}{0.66\columnwidth}
		\centering
		\includegraphics[width=0.66\columnwidth]{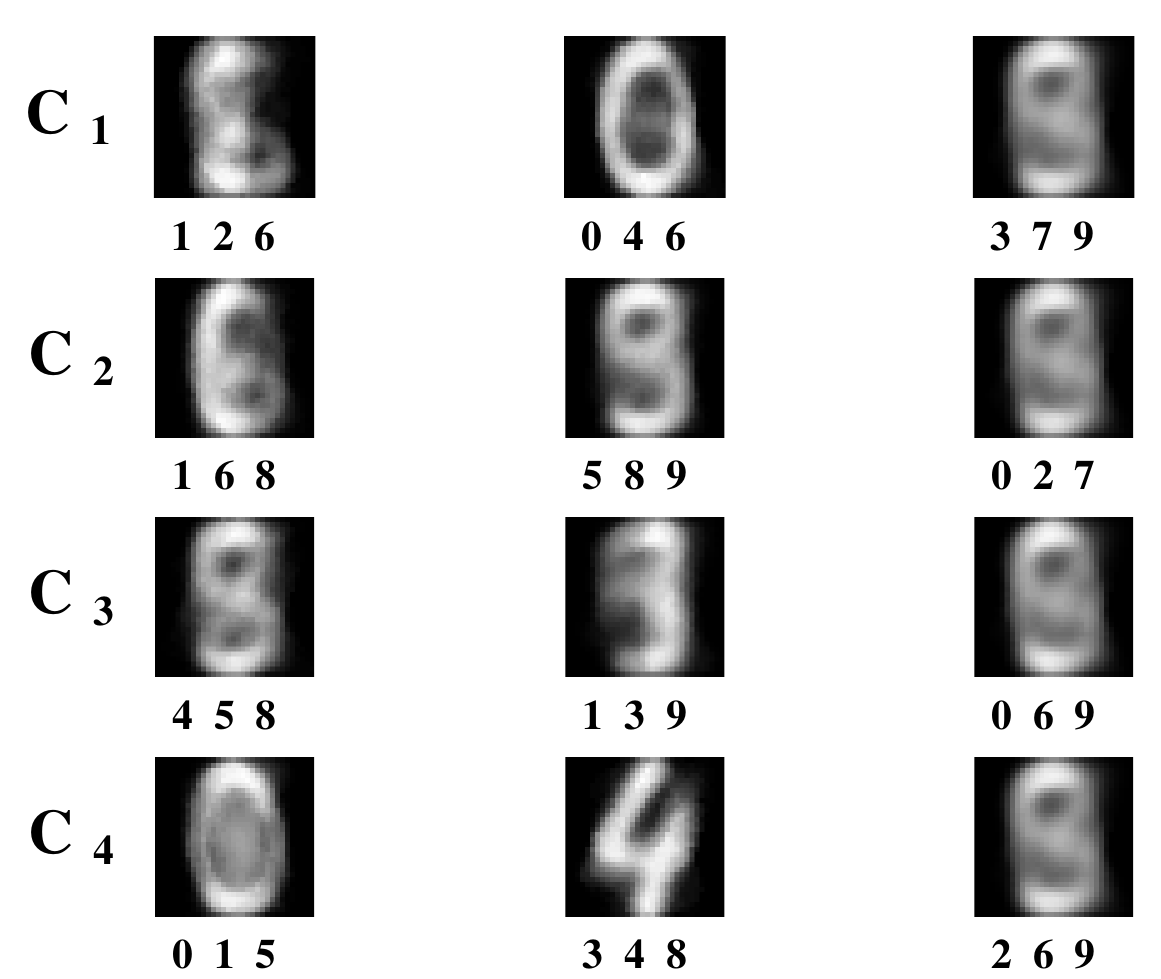}
		\caption{DMClusts}
	\end{subfigure}
	\quad
	\begin{subfigure}{0.66\columnwidth}
		\centering
		\includegraphics[width=0.66\columnwidth]{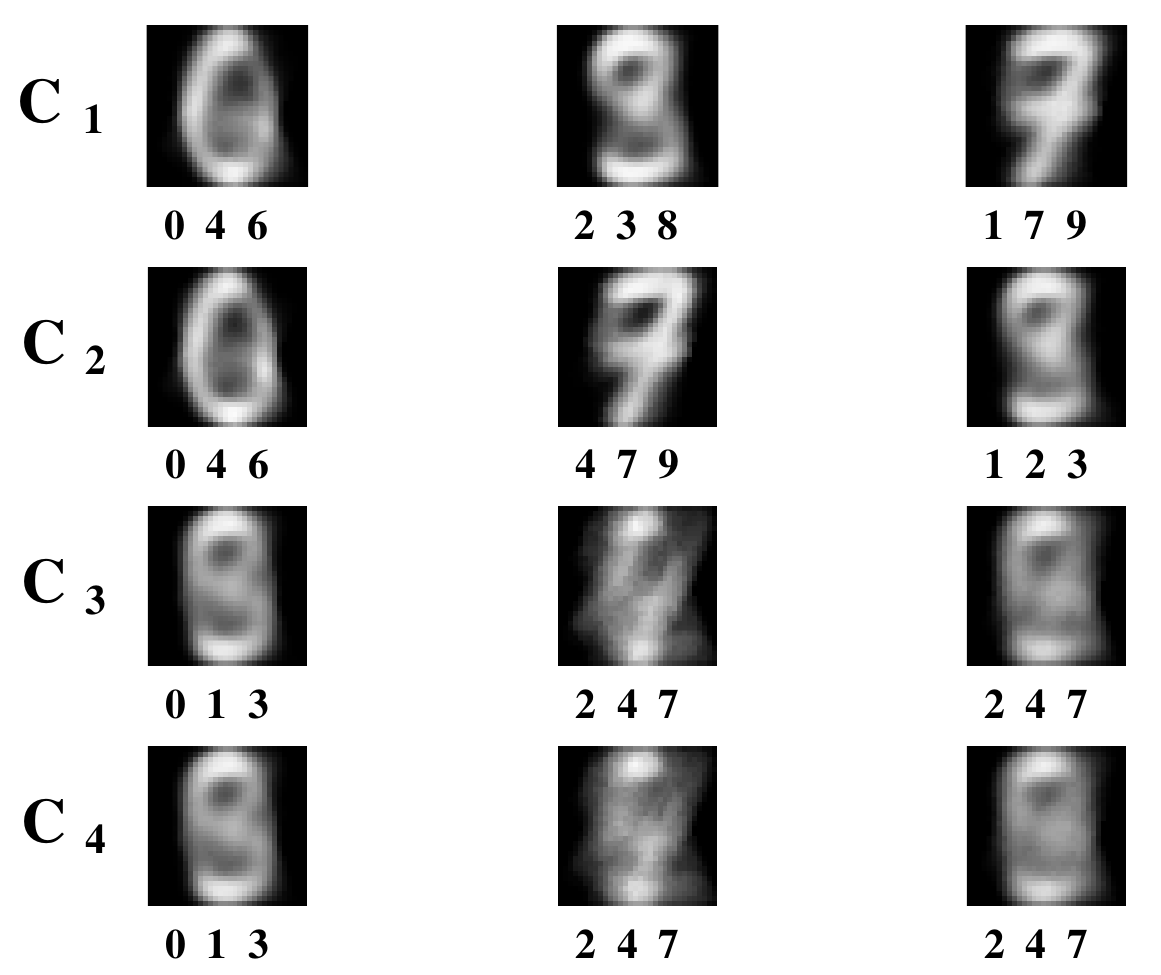}
		\caption{DMF}
	\end{subfigure}
    \quad
	\begin{subfigure}{0.66\columnwidth}
		\centering
		\includegraphics[width=0.66\columnwidth]{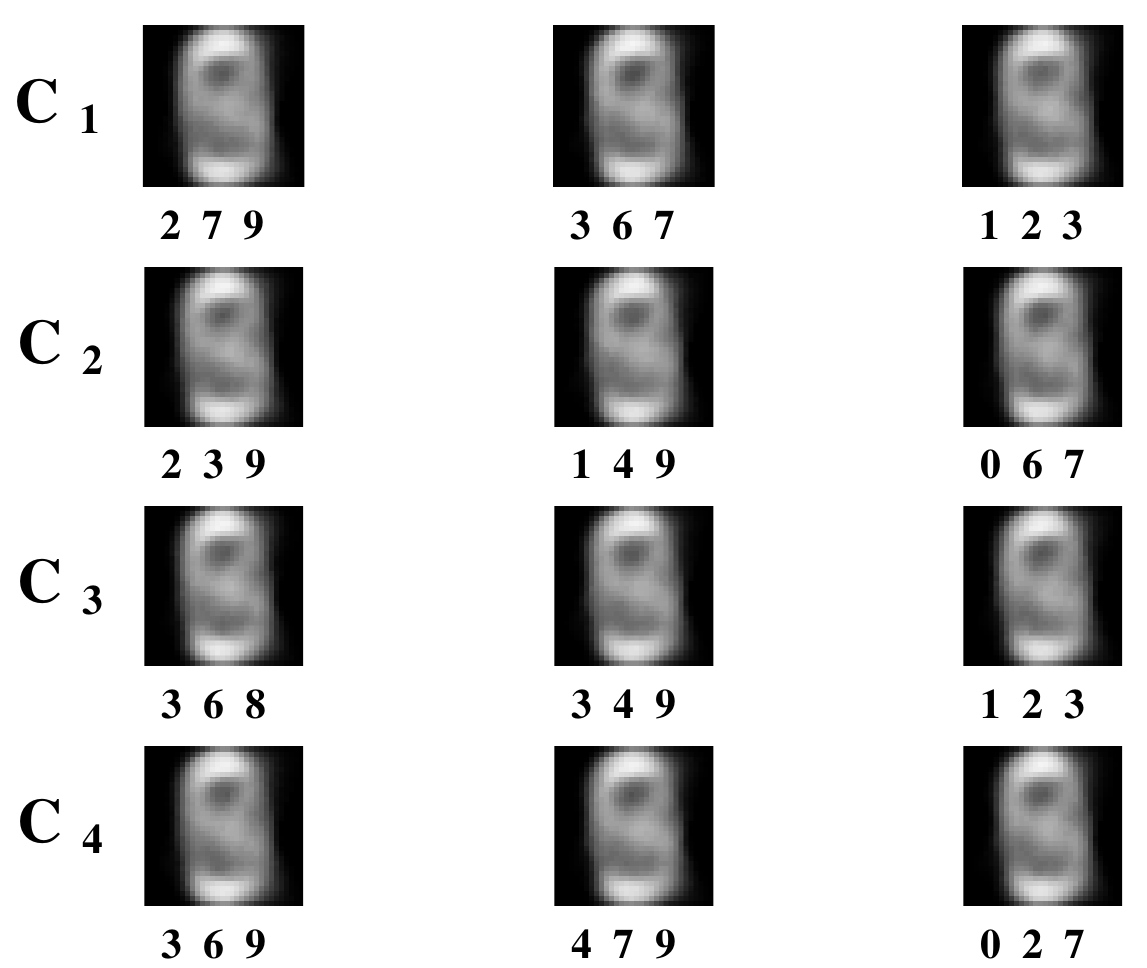}
		\caption{MVMC}
	\end{subfigure}
	\caption{Four alternative clusterings ($\mathcal{C}_1$ to $\mathcal{C}_4$) generated by DMClusts (a), DMF (b) and MVMC (c).}\label{fig:1}
    \label{fig2}
\end{figure*}

To further study whether DMClusts can generate $M\geq 3$ clusterings, we fix the number of target clusterings to $M=4$ and the number clusters for each clustering to $k=3$. Next, we apply DMClusts, DMF, and MVMC on the Handwritten dataset with images in 10 digits and visualize their clusterings in Figure \ref{fig2}. Each row of the subfigure represents a clustering and each image corresponds to the mean of the cluster. The numbers under each image are the dominant digits (not all) in the cluster. It is well known that the handwritten 10 digits are ambiguous and resemble different numbers (7 alike 4 and 3; 9 alike 5 and 7). As such, there is a tendency to group them together in different alternative clusterings. Due to the use of diversity control, DMClusts presents four clusterings without any completely overlapping clusters. In contrast, DMF does not account for  diversity and generates some largely overlapping clusters (i.e., \{0, 1, 3\}, \{2, 4, 7\} in $\mathcal{C}_3$ and  $\mathcal{C}_4$). Although MVMC also quantifies the redundancy of two objects often grouped into the same cluster of different clusterings, it still generates a heavily overlapping cluster \{1, 2, 3\} in $\mathcal{C}_1$ and  $\mathcal{C}_3$. This visual example not only confirms the effectiveness of DMClusts in generating multiple diverse clusterings, but also proves the effectiveness of our balanced redundancy quantification.

%
\begin{figure*}	
	\centering
	\begin{subfigure}{0.66\columnwidth}
		\centering
		\includegraphics[width=0.66\columnwidth]{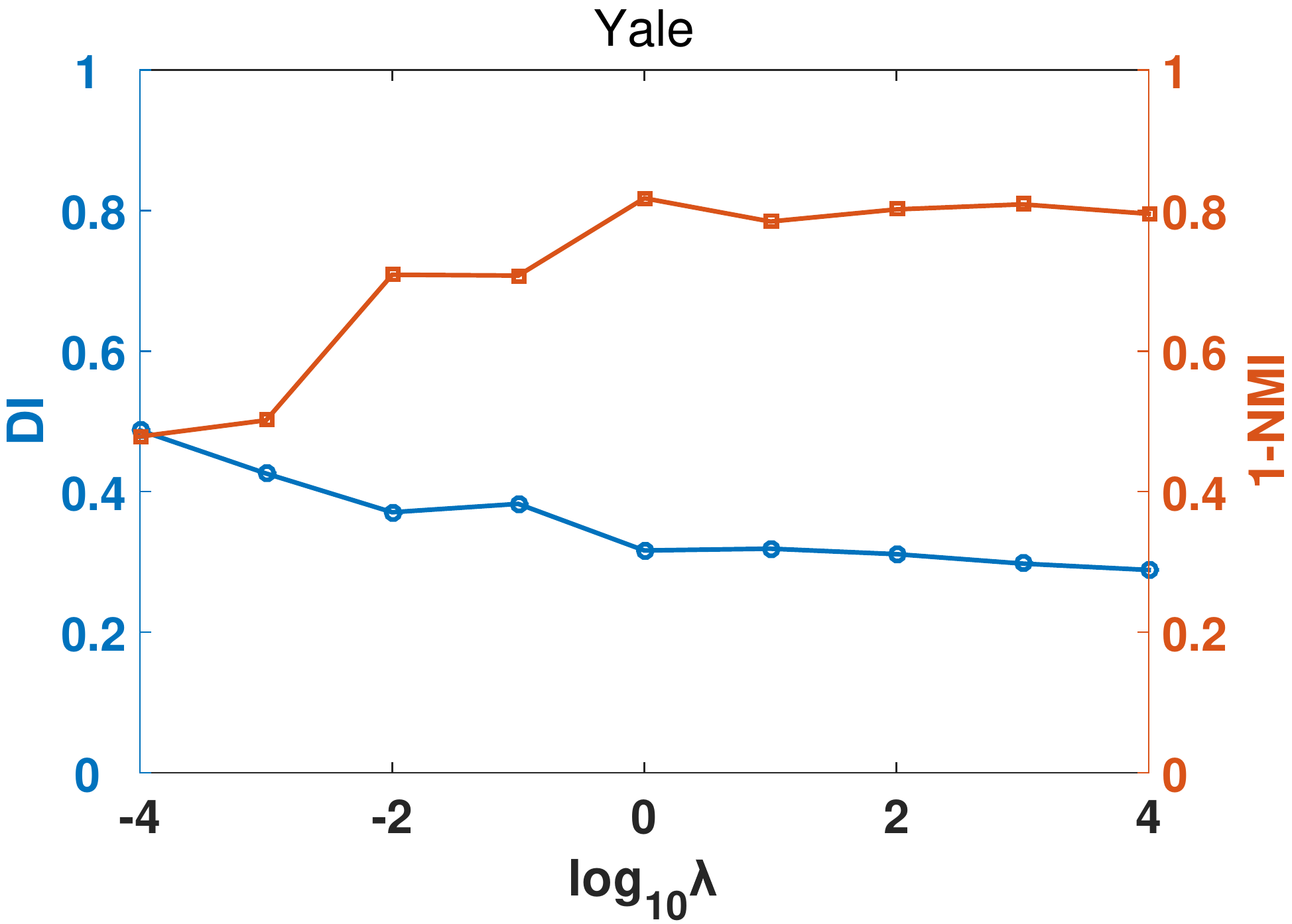}
		\caption{DI and (1-NMI) vs. $\lambda$}\label{fig3a}
	\end{subfigure}
	\quad
	\begin{subfigure}{0.66\columnwidth}
		\centering
		\includegraphics[width=0.66\columnwidth]{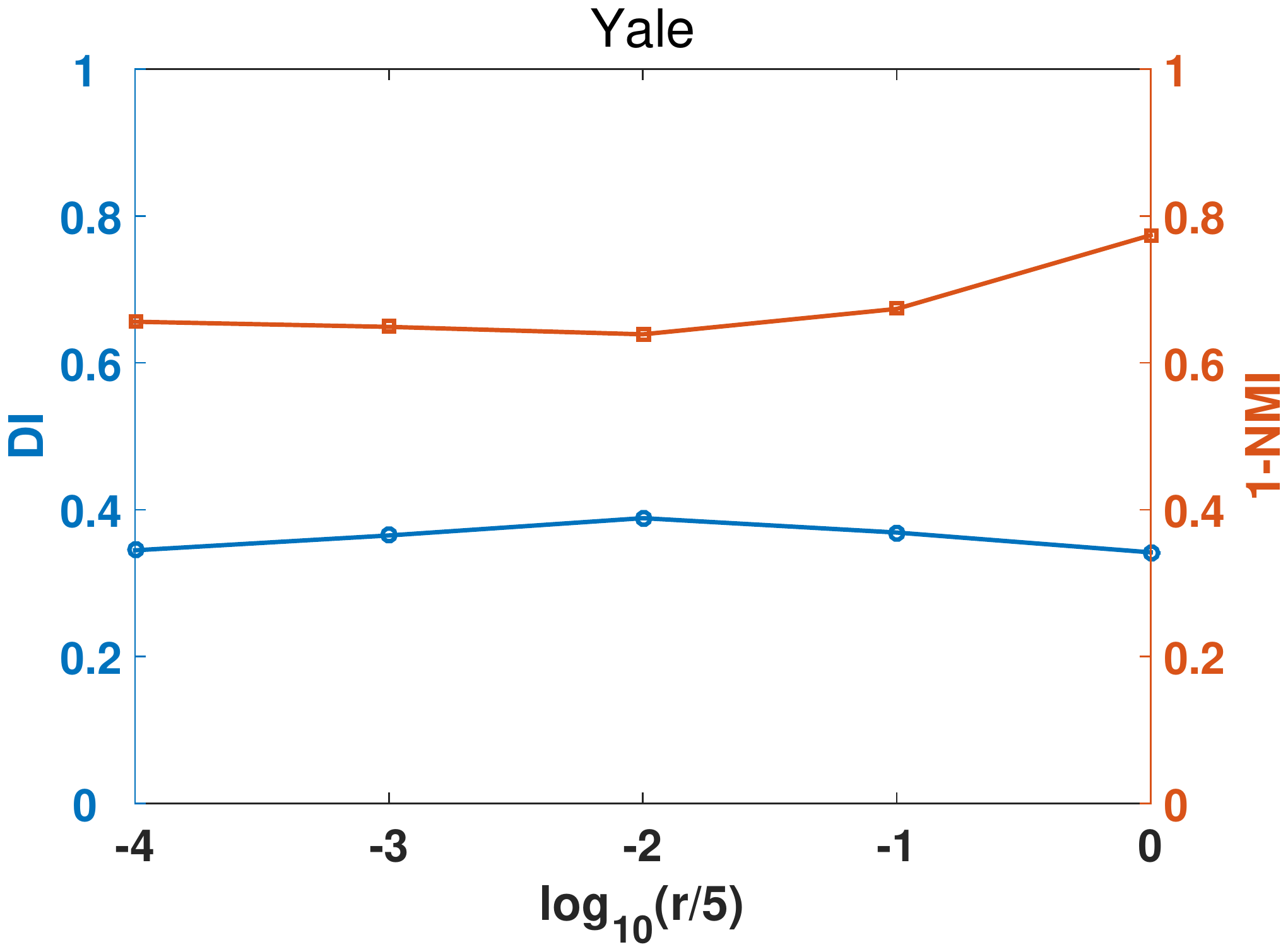}
		\caption{DI and (1-NMI) vs. $r$}\label{fig3b}
	\end{subfigure}
    \quad
	\begin{subfigure}{0.66\columnwidth}
		\centering
		\includegraphics[width=0.66\columnwidth]{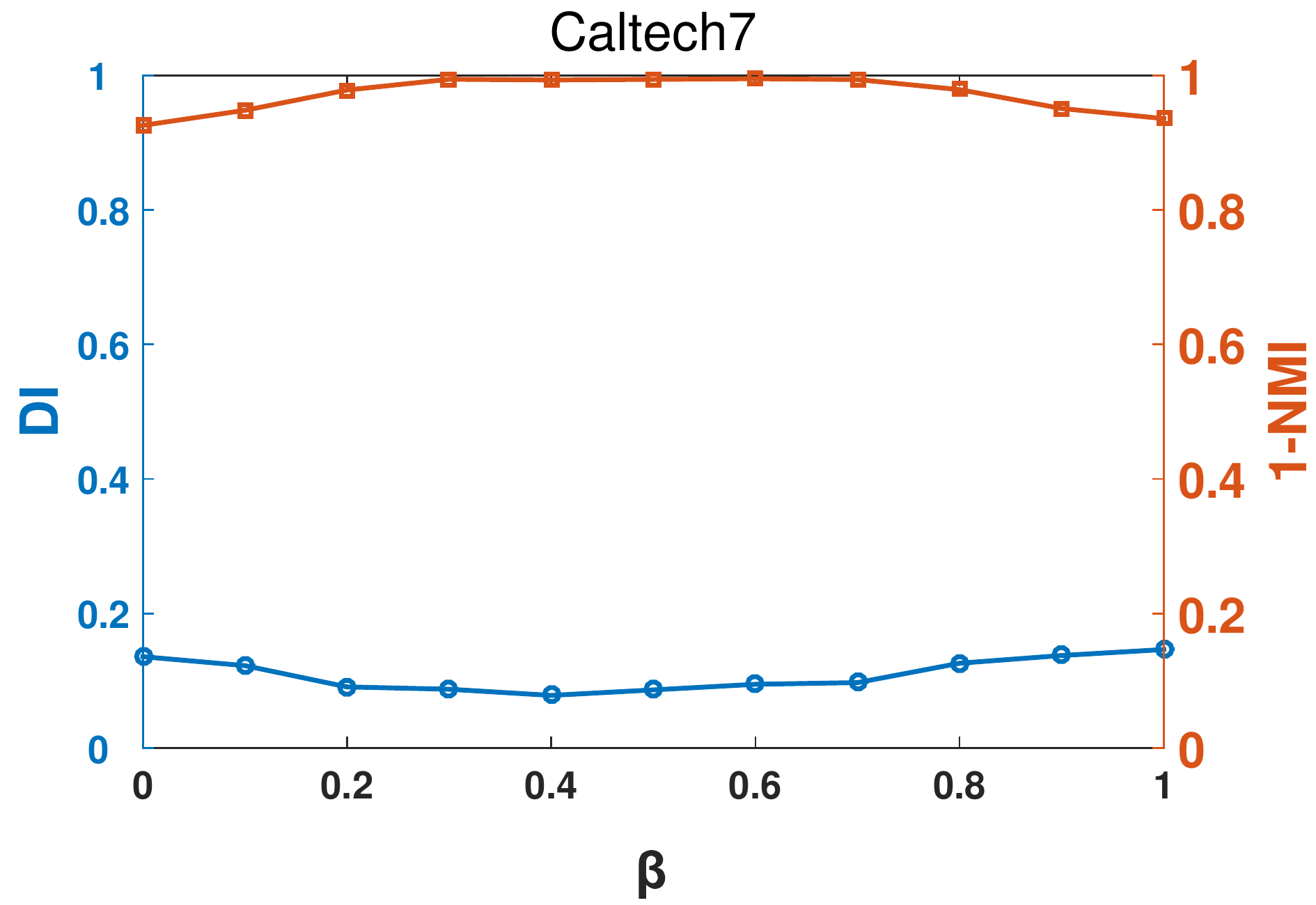}
		\caption{DI and (1-NMI) vs. $\beta$}\label{fig3c}
	\end{subfigure}
	\caption{Quality (DI) and Diversity (1-NMI) of DMClusts vs. $\lambda$, $r$ and $\beta$.}
\end{figure*}

\subsection{Parameter analysis}
Several input parameters ($\lambda$,  $r$ and $\beta$) may affect the performance of DMClusts. $\lambda$ balances the importance of deep matrix factorization and the diversity control term, $r$ controls the weight distribution assigned to input views, $\beta$ balances the redundancy of two objects placed into the same clusters and redundancy of two objects placed into different clusters of two clusterings.

We study the impact of $\lambda$ by varying it from $10^{-4}$ to $10^4$, and plot the change of Quality (DI) and Diversity (1-NMI, the larger the better) of DMClusts on the Yale dataset in Figure \ref{fig3a} with $r=0.5$, $\beta=0.4$. We find that: (i) diversity (1-NMI) steadily increases at  first but not so when $\lambda\geq 1$; (ii) the quality (DI) gradually decreases as  $\lambda$ increases, and becomes relatively stable after $\lambda \geq 1$. This pattern is explainable, since a larger $\lambda$ forces DMClusts to focus more on the diversity between clusterings, and thus may drag down the quality of the respective clusterings. Overall, this observation confirms the dilemma between diversity and quality of multiple clusterings, and shows the necessity of introducing $\lambda$ to control the redundancy.

We investigate the impact of $r$ by varying it in the grid of $\{5\times 10^{-4}, 5\times 10^{-3}, \cdots,5\}$, and report the quality and diversity  of DMClusts on the Yale dataset in Figure \ref{fig3b} with $\lambda=0.01$, $\beta=0.4$. The quality slightly rises as $r$ increases, and the diversity remains stable. When $r>0.05$, the diversity steadily increases and the quality gradually decreases, due to the increased diversity and the known trade-off between diversity and quality. This is because a too small $r$ gives nearly equal weights to all the views, while a moderate $r$ can assign different sets of weights to these views, which helps to generate diverse clusterings, as exampled in Figure \ref{fig:case}.

To study the benefit of our balanced redundancy quantification term, we vary $\beta$ from 0 to 1 and report the results in Figure \ref{fig3c} with $r=0.5$, $\lambda=0.01$. We observe that the diversity (1-NMI) increases as $\beta$ increase but turns to reduce as $\beta>0.7$. Due to the dilemma between quality and diversity, the quality shows a reverse trend. Neither $\beta=1$ nor $\beta=0$ gives the highest diversity, and $\beta \in [0.3, 0.7]$ gives an NMI$\approx$0. This observation proves the contribution of considering the previously overlooked redundancy due to two samples placed in different clusters of two clusterings, and also justifies the effectiveness of our balanced redundancy quantification term. In addition, it clarifies why our DMClusts obtains a better diversity between clusterings. We observe that $\beta=1$ (NMI: 0.064) gives a larger diversity (by $\approx$16\%) than $\beta=0$ (NMI: 0.075). This suggests the redundancy two samples in the same clusters of two clusterings is more important than the redundancy they in different clusters. Overall, these two types of redundancy complement each other and help to generate multiple clusterings with improved diversity.

We further study the impact of the number of clusters, layer size $K_l$ and of different $M$. We also make a runtime experiment and show that DMClusts not only outperforms the state-of-the-art methods in exploring multiple clusterings with quality and diversity, but also holds a moderate efficiency. The results and analysis as well as convergence analysis can be found in the Supplementary file.
Finally, we want to remark that, all the four metrics do not depend on the ground-truth labels of the tested dataset, so the suitable values for  parameters can be chosen based on the user's preference toward quality or diversity.

\section{Conclusion}
In this paper, we introduce DMClusts to explore multiple clusterings from multi-view data, which is an interesting, practical  but overlooked clustering topic that conjoins multi-view clusterings and multiple clusterings. DMClusts adapts the deep matrix factorization to a deep learning approach, and introduces a novel balanced diversity quantification term to seek multiple diverse clusterings of quality. DMClusts shows a superior effectiveness and efficiency than state-of-the-art competitive solutions. We will investigate a principle to determine a suitable number of layers (clusterings).


\section{Acknowledgments}
This work is supported by NSFC (61872300 and 61873214), Fundamental Research
Funds for the Central Universities (XDJK2019B024), Natural Science Foundation of CQ CSTC
(cstc2018jcyjAX0228) and by the King Abdullah University
of Science and Technology (KAUST), Saudi Arabia. The code and Supplementary file of DMClusts is available at http://mlda.swu.edu.cn/codes.php?name=DMClusts.

\bigskip
\bibliographystyle{aaai}
\bibliography{DMF_Bib}
\end{document}